# Edge-aware Hard Clustering Graph Pooling for Brain Imaging

Cheng Zhu[# 1], Jiayi Zhu[# 2], Xi Wu[1], Lijuan Zhang[1], Shuqi Yang[1], Ping Liang[1], Honghan Chen[1], Ying Tan[* 1]

[1] The Key Laboratory for Computer Systems of State Ethnic Affairs Commission, Southwest Minzu University, Chengdu, 610041, China
[2] State Key Laboratory of Cognitive Neuroscience and Learning, Beijing Normal University, Beijing 100875, China
190852112021@stu.swun.edu.cn, zhujiayii0102@163.com, 220854112016@stu.swun.edu.cn, 13769472963@163.com, ysqwymail@163.com, 21700059@swun.edu.cn, honghanchen@hotmail.com, ty7499@swun.edu.cn

## Abstract

Graph Convolutional Networks (GCNs) can capture non-Euclidean spatial dependence between different brain regions. The graph pooling operator, a crucial element of GCNs, enhances the representation learning capability and facilitates the acquisition of abnormal brain maps. However, most existing research designs graph pooling operators solely from the perspective of nodes while disregarding the original edge features. This confines graph pooling application scenarios and diminishes its ability to capture critical substructures. In this paper, we propose a novel edge-aware hard clustering graph pool (EHCPool), which is tailored to dominant edge features and redefines the clustering process. EHCPool initially introduced the 'Edge-to-Node' score criterion which utilized edge information to evaluate the significance of nodes. An innovative Iteration n-top strategy was then developed, guided by edge scores, to adaptively learn sparse hard clustering assignments for graphs. Additionally, a N-E Aggregation strategy is designed to aggregate node and edge features in each independent subgraph. Extensive experiments on the multi-site public datasets demonstrate the superiority and robustness of the proposed model. More notably, EHCPool has the potential to probe different types of dysfunctional brain networks from a data-driven perspective. Method code: https://github.com/swfen/EHCPool

## Introduction

The brain can be regarded as a complex graph structure in non-Euclidean space. In recent years, many studies have utilized graph convolution kernels to investigate the brain since graph convolutional networks (GCNs) are among the few advanced deep learning methods that can directly analyze graph structures (Jiang et al., 2020) (Dsouza et al., 2021) (Wen et al., 2022). However, in graph-level classification tasks, an elaborate graph pooling operator occupies an indispensable role in learning abstract representations of the brain. Since, graph pooling, which is tailored to the unique characteristics of brain image data, remains a relatively unexplored area of study. Therefore, it is crucial to design advanced graph pooling operators based on specific data scenarios.

---

[#] These authors contributed equally to this work
[*] Corresponding Author

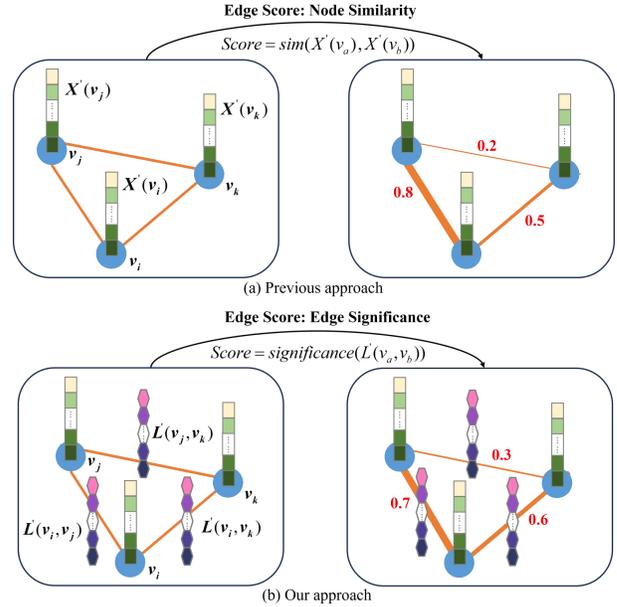

Figure 1. Comparison of Edge Score Calculation. The edge scores obtained using the previous approach (a) are derived from the similarity of neighboring nodes, while our approach (b) generates edge scores based on the significance of the edge features.

Existing graph pooling methods mostly disregard the significance of edge features. As depicted in Figure 1, while certain approaches may prioritize edges, their primary focus is typically on edge weight values determined by neighboring node similarity rather than scores related to edge features themselves. e.g., EdgePool (Diehl 2019), MincutPool (Bianchi, Grattarola, and Alippi 2020), and Co-Pool (Zhou, Yin, and Tsang 2023). However, a similar situation exists in all types of complex systems and graph problems (e.g., brain image analysis, path selection, and recommendation system), where the significance or utilization of edge information may surpass than that of node information.

Take the brain imaging problem as an example, functional Magnetic Resonance Imaging (fMRI) is the mainstream detection technology used to explore the brain functional activities (Huettel 2012; Lin et al.2016). Functional connectivity (FC) is widely recognized as a crucial feature in brain research (Bifone, Gozzi, and Schwarz 2010; Nili et al. 2023). FC measures the correlation of activities in different brain regions and is an objective means for studying brain activity. The graph structure implements brain regions as nodes and FC as edges between those nodes, aligning with the logical structure of brain data and has been recognized by previous studies (Li et al. 2019; Li et al. 2021). However, due to insufficient and unreliable methods for evaluating edge information in graphs, several fMRI studies have utilized GCNs by directly assigning logical edge information (i.e., FC) to the graph nodes, which neglects the design of edge information (Venkatapathy et al. 2023; Klepl et al. 2022; Pitsik et al. 2023). This approach results in the loss of structural representations of the graphs. Thus, graph pooling operators targeted at prominent edge features would provide a means to surmount current technical obstacles and attain superior model performance.

On the other hand, Graph clustering method in GCN facilitates capture of abnormal networks in the brain. But most of the emerging graph clustering methods have used soft clustering (a node belongs to more than one cluster). For example, AHDDC (Dong et al. 2023) leverages an attention-based soft clustering method to reinforce the structural information. AGCC (He et al. 2022) is an end-to-end parallel adaptive soft clustering model that disseminates optimal data representations. Learnable Pooling (Gopinath et al. 2022) proposed a soft clustering approach that learn the intrinsic aggregation of nodes based on graph embedding. Most methods use clustering to classify nodes into different subgraphs (clusters) and subsequently aggregate the information from each subgraph into corresponding basis vectors for training to exclude data redundancy or aggregate high-value feature information. However, utilizing soft clustering approaches may lead to the repeated focus on individual nodes with high-value (highly abnormal) information by different subgraph basis vectors. This may hinder the model's adaptability in discovering critical subgraphs with differentiation and ultimately decrease its interpretability (Zhang et al. 2023; Wagner et al. 2022).

To overcome the limitations of existing graph pooling operators as well as to adapt the brain fMRI data characteristics, we propose a clustering graph pooling method that supports edge features called Edge-aware hard clustering graph pooling (EHCPool). Detailedly, EHCPool initially presents a method for calculating scores of edge features, assessing their significance in an adaptive way. Then, the Edge-to-Node criterion is utilized for determining the target node score based on the importance of neighboring edges. Thereafter, based on the score ranking of all nodes, the most informative node is selected as the core node (cluster center) of the current subgraph (cluster) in order using the Iteration n-top strategy. And for each core node, a finite number of neighborhood nodes with maximum neighboring edge scores are selected to form the corresponding subgraphs. Finally, the N-E aggregation aggregates the subgraph vicinity information to the core node, which in turn forms the basis vector representing each subgraph for the output.

In summary, EHCPool aims to revolutionize the graph clustering process by leveraging dominant edge features as a guide. The principal contributions can be outlined as follows:

- The feature evaluation criterion is proposed for measuring the nodes significance based on edge feature information scores: Edge-to-Node. This strategy establishes a linkage between node and edge feature information, and holistically considers both global and local representations.
- A subgraph acquisition strategy is presented utilizing adaptive sparse hard clustering: Iteration n-top. This strategy aims to prevent the duplication of high-value information representations across subgraphs and has the potential to enhance model interpretability.
- A targeted aggregation strategy is designed for fusing node and edge feature information: N-E aggregation. This strategy introduces new ideas for information aggregation within graph structures that include both node and edge features.
- To the best of our knowledge, we are the first that proposed a graph pooling method that accommodates edge features. It provides a novel application paradigm pertaining to GCNs data challenges, where the significance of edge features exceeds that of node features.

## Related Work

**Graph Pooling Method.** From the processing strategy of graph feature information, the current mainstream hierarchical graph pooling methods can be divided into three types. (i) Sparse selection pooling. This strategy scores all node features and directly retains the most important few nodes. Because no node aggregation is performed, although sparsity is better guaranteed, it is not conducive to preserving the node information of the entire graph. e.g., Top-kPool (Gao and Ji 2019), SAGPool (Lee, Lee, and Kang 2019), etc. (ii) Dense clustering pooling. This method globally assigns all nodes to a fixed number of clusters by soft clustering, and then obtains the basis vector of each cluster. This strategy better ensures the completeness of feature information due to the basis vectors of each cluster containing all node information, but it also increases the arithmetic burden. e.g., DiffPool (Ying et al., 2018), etc. (iii) Sparse Cluster Pooling. This strategy combines the advantages of the two types mentioned by assigning a finite number of local nodes to a fixed number of different clusters, and thus obtaining the basis vectors of each cluster. This approach maintains graph information completeness while ensuring sparsity. Examples include ASAP (Ranjan, E.et al 2019), etc.

| Property | Top-k Pool | SAG Pool | Diff-Pool | ASAP | Ours |
|---|---|---|---|---|---|
| Sparse | ✓ | ✓ | | ✓ | ✓ |
| Node aggregation | | | ✓ | ✓ | ✓ |
| Graph clustering | | | ✓ | ✓ | ✓ |
| Edge feature support | | | | | ✓ |

Table 1: Properties in Graph pooling methods

However, the aforementioned methods only focus on node features, which hinders their ability to adapt to edge features during the clustering implementation process. Meanwhile, this limitation constrains the potential use cases for graph pooling methods. Therefore, this study presents a novel approach for sparse clustering pooling through the utilization of edge features as an entry point. In addition, Table 1 shows the property comparison of above graph pooling operators with EHCPool, from which it can be observed that our method consists of additional properties.

**Graph Clustering.** Recent studies of clustering for GCNs tends to focus on two primary methods: spectral methods (Ma et al. 2019) and non-spectral methods (Ying, R. et al. 2018). Since spectral techniques require significant computational resources (Ranjan, E. et al 2019) and there is a shortage of brain imaging data, which could exacerbate overfitting issues, this study designed the clustering process of EHCPool mainly from a non-spectral perspective.

In addition, the clustering methods can be categorized into two types: hard clustering and soft clustering. Compared to soft clustering, hard clustering has the benefit of being easy to comprehend and interpret, while the disadvantage is that it is unable to handle situations where there is overlap between class clusters, and it is not applicable to complex data structures (Horta and Campello 2015). However, in mechanism studies of functional brain network, all network parcellations have been derived from hard clustering algorithms (Ji et al., 2019; Yeo et al., 2011), and the parcellated brain regions do not overlap. Meanwhile, parcellation templates generally divide the brain into tens to hundreds of regions, which is not a very complex graph structure. Based on the current state of research in this field, the hard clustering algorithm shows more promise and generalizability for application.

Therefore, this study focuses on developing deep learning tools that have the potential to analyze various types of functional brain networks using hard clustering techniques.

## Method

**Graph Definition and Feature Construction.** FC is divided into static FC (sFC) and dynamic FC (dFC). dFC provides valuable information regarding the organization of functional brain networks that cannot be obtained by sFC (Díez-Cirarda et al. 2018), and dFC is more strongly associated with psychiatric disorder dysfunction (Espinoza et al. 2019). Therefore, dFC was selected in this study to describe the dynamic functional activities of the brain. Meanwhile, DSF-BrainNet (Zhu et al. 2023) was selected as the graph structure input for the GCN model.

DSF-BrainNet comprises two parts: node and edge features. Features are organized as time-series data, and both of which maintain the synchronization property. The fundamental notation is defined as follows: $\mathcal{G} = (\mathcal{V}, \mathcal{E})$ denotes a graph, where $\mathcal{V} = \{v_1, v_2, \cdots, v_n\}$ denotes the set of brain region nodes and $n$ denotes the number of nodes. Similarly, $\mathcal{E} = \{(v_i, v_j)\}$ denotes the set of edges between graph nodes, where $\mathcal{E}$ contains $M$ elements. The function $X : \mathcal{V} \mapsto \mathbb{R}^{d_v}$ assigns features to each node, and the function $L : \mathcal{E} \mapsto \mathbb{R}^{d_e}$ assigns features to each edge. Thus, $X(v_i), i = 1, 2, \cdots, n$ is the feature vector of node $v_i$; $L(v_i, v_j)$ is the edge feature vector between nodes $v_i, v_j$.

For a node $v_i$, the set of its neighboring nodes is defined as $Ne(v_i)$, that is, $Ne(v_i) = \{v_j | (v_i, v_j) \in \mathcal{E}\}$, where $v_j$ denotes the neighboring nodes of the $v_i$. In this study, it should be noted that the neighboring information of each node is stored in a matrix defined as $A \in \mathbb{R}^{2 \times M}$, where 2 indicates that $A$ has two feature dimensions: the first dimension is the index $i$ of the initial node $v_i$, and the second is the index $j$ of neighboring node $v_j$. M denotes the total number of edges in $\mathcal{G} = (\mathcal{V}, \mathcal{E})$. The specific computational procedure for node and edge feature construction is detailed in Supplement (section A).

**Graph Convolution Module.** This study used Edge-Conditioned Convolution (ECC) (Simonovsky and Komodakis 2017) as the graph convolution module. Distinguishing from other traditional graph convolution methods, ECC is a generalized method that supports edge feature design, which is defined as follows:

$$X^l(v_i) = \frac{1}{|Ne(v_i)|} \sum_{v_j \in Ne(v_i)} F^l(L(v_i, v_j); w^l) X^{l-1}(v_j) + b^l \quad (1)$$

where $l$ is the layer index in the feedforward neural network, $b^l \in \mathbb{R}^{d_l}$ is a learnable bias and function $F^l(\cdot)$ is a customized multilayer perceptron (MLP) that parameterized by learnable network weights $w^l$.

## EHCPool: Proposed Method

**Overall Architecture.** To enhance the focus on edge feature information during graph pooling, a hard clustering pooling method named EHCPool is proposed. As shown in Figure. 2, EHCPool incorporates three novel sub-methods at its core:

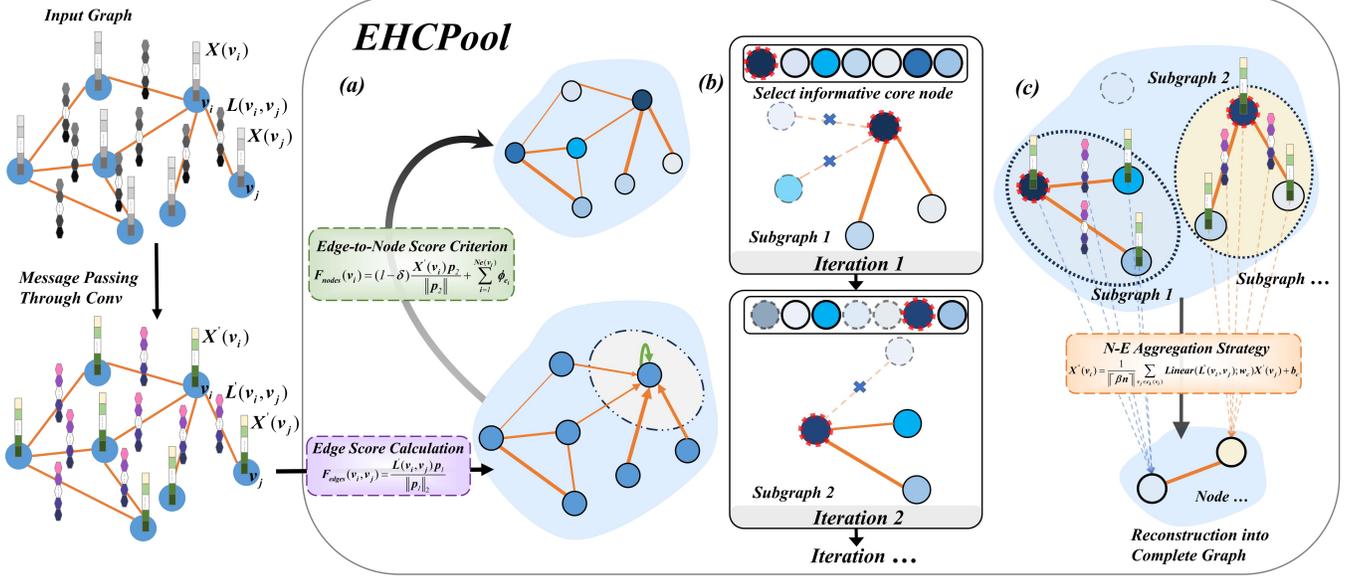

Figure 2: Overall EHCPool Architecture. The input graph is constructed with multidimensional node and edge features. It is then fed into the EHCPool for further representation learning after message passing through the convolutional layer. (a) Score calculation using edge feature information. The white dashed box displays the specific flow of the Edge-to-Node criterion, where edge scores are depicted by edge thickness and node scores are depicted by the node color shade. (b) Cluster formation using an iterative hard clustering strategy. The node outlined represents the most informative node within the current node set, i.e., the core node within the subgraph. Regular nodes corresponding to selected core nodes are chosen based on edge information scores. (c) Aggregate the core node vicinity information in each subgraph. The comprehensive information is then aggregated to forms the basis vectors of each subgraph, which in turn create the EHCPool output. Finally, the output nodes connect to form a complete graph that serves as input for the next layer of the network.

(a) feature score criterion using 'Edge-to-Node'; (b) hard cluster formation using Iteration n-top; and (c) information aggregation using N-E aggregation.

**Edge-to-Node score criterion.** Existing graph pooling methods typically induce a coarser subgraph from the node perspective, but this overlooks the critical role of edge features. Furthermore, when dealing with data problems where edge feature information surpasses the importance of nodes, emphasizing the value of edge information during graph pooling is justifiable. Thus, in EHCPool, a strategy for calculating the edge feature information score was initially presented to determine the significance of the edge information, in which the edge feature information score $\phi_{e_t}, t=1,2,\cdots,M$ is expressed by the function $F_{edges}(v_i, v_j)$, and the set of all edge feature information scores is denoted by $\phi_{edges} \in \mathbb{R}^M$. The implemented is as follows:

$$F_{edges}(v_i, v_j) = \frac{L'(v_i, v_j) p_1}{\|p_1\|_2} \quad (2)$$

$$\phi_{edges} = [\phi_{e_1}, \phi_{e_2}, \ldots, \phi_{e_M}]$$

(3)

$$L^{*}(v_i, v_j) = L'(v_i, v_j) \odot \sigma(F_{edges}(v_i, v_j)) \quad (4)$$

where $L'(v_i, v_j) \in \mathbb{R}^{1 \times d_e}$ is obtained from the edge features after message passing by the convolution module, $p_1 \in \mathbb{R}^{d \times 1}$ denotes the learnable projection vector. $\|\cdot\|_2$ denotes the Euclidean distance ($L_2$ norm). $L^{*}(v_i, v_j) \in \mathbb{R}^{1 \times d_e}$ denotes the edge features after the scores are updated, '$\odot$' is the (broadcasted) element-wise multiplication, $\sigma(\cdot)$ represents the sigmoid activation function. Subsequently, we define a node feature information score criterion named 'Edge-to-Node'. Noteworthy is that this strategy takes into account the significance of edge details while also acknowledging node information. Specifically, it involves assessing the score of targeted node and all of its neighboring edges, which is an evaluation criterion that comprehensively considers both global and local representations. Likewise, the node feature information score $\phi_{v_i}$ is expressed by the function $F_{nodes}(v_i)$, and the node feature information score set is expressed by $\phi_{nodes} \in \mathbb{R}^n$, executed as follows:

$$F_{nodes}(v_i) = (1-\delta)\frac{X'(v_i)p_2}{\|p_2\|_2} + \sum_{t=1}^{Ne(v_i)} \phi_{e_t} \quad (5)$$

$$\phi_{nodes} = [\phi_{v_1}, \phi_{v_2}, ..., \phi_{v_n}] \quad (6)$$

where $\delta \in [0,1]$ serves as the moderating factor, controlling the degree of node information neglect. Larger $\delta$ values indicate less emphasis placed on node scores, resulting in less localized information included (generally defaults to 0). $X'(v_i) \in \mathbb{R}^{1 \times d_v}$ is obtained by passing the node features through the convolution module. $p_2 \in \mathbb{R}^{d_v \times 1}$ represents the learnable projection vector.

**Iteration n-top cluster strategy**. Iteration n-top is a hard assignment with sparsity, which effectively reduces the computational complexity. Prior to initiating clustering, two hyperparameters, $\gamma$ and $\beta$, need to be predetermined: $\gamma$ controls the number of subgraphs (clusters) to be retained, and $\beta$ controls the ratio of nodes maintained in each subgraph. Considering one of the iterative processes as an example, the cluster center (or core node) of the subgraph (cluster) is identified as the most informative node based on the node feature information scores, as shown in Eq. (7):

$$idx_{max} = \arg\max_{v_i \in \mathcal{V}}(\phi_{nodes}) \quad (7)$$

where the function $\arg\max(\cdot)$ is used to retrieve the maximum value index $idx_{max}$ in the set. Therefore, the index locates the core node $v_{idx_{max}}$ in the subgraph, denoted as $v_{Cnode}$. At the same time, we can fetches the neighbors' edge indexes of this node from the adjacency matrix $A$, denoted as $idx_{Nedges}$. Next, the neighboring edges information scores are sorted, and subsequently, the neighboring nodes corresponding to the top $(\lceil \beta n \rceil - 1)$ scoring edges are selected as the regular nodes for the subgraph. The edges from the regular nodes to the core nodes are also retained. By employing a judging criterion oriented towards edge information scores, a more comprehensive assessment of the contributions of nodes and edges within the graph structure is achieved. The formulas are as follows:

$$idx_{Sortedges} = top\_rank(\phi_{idx_{Nedges}}, (\lceil \beta n \rceil - 1)) \quad (8)$$

$$idx_{Nenodes} = search\_A(idx_{Sortedges}) \quad (9)$$

$$\mathcal{V}_{SubNodes} = \mathcal{V}_{idx_{Nenodes}}, \quad \mathcal{E}_{SubEdges} = \mathcal{E}_{idx_{Sortedges}} \quad (10)$$

where $top\_rank(\cdot)$ sorts the neighboring edges based on their edge scores and returns the top $(\lceil \beta n \rceil - 1)$ indexes. $search\_A(\cdot)$ utilizes the chosen neighboring edge indexes $idx_{Sortedges}$ to acquire a set of neighboring node indexes that correspond to the given neighboring edges in the adjacency matrix $A$, denoted as $idx_{Nenodes}$. $\mathcal{V}_{SubNodes} \subset \mathcal{V}$, which is a subset of $\mathcal{V}$, contains regular nodes in the subgraph. $\mathcal{E}_{SubEdges} \subset \mathcal{E}$, which is a subset of $\mathcal{E}$, contains neighboring edges in the subgraph that are connected to regular nodes. Therefore, we define the subgraph dominated by the core node $v_i$ as $c(v_i) = \{v_j; (v_i, v_j) \in \mathcal{E}_{SubEdges}\} \cup \{v_i\}$. Finally, the updated node score set $\phi_{update\_nodes}$ is obtained by removing the selected nodes from the initial set of node feature information scores:

$$idx_{sub} = idx_{max} \bigcup idx_{Nenodes} \quad (11)$$

$$Z_{mask} = \Gamma_{idx_{sub}} \quad (12)$$

$$\phi_{update\_nodes} = \phi_{nodes} \odot Z_{mask} \quad (13)$$

where the symbol ' $\bigcup$ ' denotes the union operation, $\Gamma_{idx_{sub}}$ denotes the index operation, and $Z_{mask}$ denotes the feature mask (mask as 0). Hard cluster formation is completed, and the process is iterated until $\gamma$ subgraphs $c_k(v_i), k = 1, 2, ..., \gamma$ are obtained.

**N-E Aggregation strategy.** To enhance the representation learning of graph structures with both node and edge features, this study introduces N-E Aggregation, a novel information aggregation strategy. This technique performs independent information aggregation within each subgraph, allowing it to aggregate vicinity information, including nodes and edges, to the core nodes. As a result, it overcomes the limitations of traditional graph pooling, which relies solely on node updates and loses structural information. It ensures that information interactions between subgraphs are isolated, avoiding redundancy and confusion. After completing the aforementioned hard clustering process, the objective is to acquire the subgraph representation by consolidating the regular nodes information $v_j \in c_k(v_i)$ and the edges neighboring each subgraph with its core node $v_i$. The aggregation process is as follows:

$$X''(v_i) = \sum_{v_j \in c_k(v_i)} Linear(L'(v_i, v_j); w_c) X'(v_j) + b_c \quad (14)$$

where $X''(v_i) \in \mathbb{R}^{1 \times d_v}$ is the basis vector representing subgraph $c_k(v_i)$ after information aggregation. $w_c \in \mathbb{R}^{(d_v \times d_v) \times d_e}$ and $b_c \in \mathbb{R}^{1 \times d_v}$ denote the trainable weights and biases, respectively. $Linear(\cdot)$ transforms edge feature vectors with the input dimensions of $\mathbb{R}^{1 \times d_e}$ into a weight matrix with the output dimensions of $\mathbb{R}^{d_v \times d_v}$. Subsequently, this weight matrix is matrix-multiplied by the node feature vector $X'(v_j)$ to obtain the information passed to core node $v_i$ from the neighboring edges $L'(v_i, v_j)$ and the corresponding neighboring node $v_j$. The EHCPool algorithm is detailed in the Supplement (section B).

# Experiments

## Experiment Setting

**Datasets and Preprocessing.** In this study, three public datasets were chosen for model evaluation: the COBRE (Mayer et al. 2013), UCLA (Bilder et al. 2020), and REST-meta-MDD (Chen et al. 2022). Among these, the REST-meta-MDD dataset was used as a multi-site mixed dataset for major depressive disorder (MDD). It contains rs-fMRI images of 200 subjects, including 100 patients with MDD and 100 healthy controls (HC). Similar to the above, we combined the COBRE and UCLA datasets to form a mixed dataset for schizophrenia (SZ) to evaluate the model's robustness. It contains rs-fMRI images of 192 subjects, including 89 patients with SZ and 103 HC. The relevant details and preprocessing procedures are described in the Supplement (section C).

**Metrics.** Four metrics were utilized to evaluate the model's performance: accuracy (ACC), sensitivity (SEN), specificity (SPE), and the F1-Score.

**Baselines.** The comparison experiments were conducted in two parts. The first part involved comparing seven state-of-the-art (SOTA) models commonly or exclusively used for brain data. The second part involved comparing seven generalized graph pooling methods. Optimal performance was achieved by tuning the hyperparameters. (i)Brain SOTA models: Support Vector Machine (SVM) (Vapnik 1998), Graph Attention Network based learning and interpreting method (GAT-LI) (Hu et al. 2021), Brain Graph Neural Network (BrainGNN) (Li et al. 2021), GCN (Kipf and Welling 2016), Convolutional Recurrent Neural Network (CRNN) (Lin et al. 2022), High-order functional connectivity network (HFCN) (Pan et al. 2022), and Sparse Convolutional Neural Network (SCNN) (Ji, Chen, and Yang 2021). (ii) Graph pooling methods: Graph U-Nets (Top-kPool) (Gao and Ji 2019), Self-Attention Graph Pooling (SAGPool) (Lee, Lee, and Kang 2019), Edge Contraction Pooling (EdgePool) (Diehl 2019), Spectral clustering Graph Pooling (MincutPool) (Bianchi, Grattarola, and Alippi 2020), Co-Pool (Zhou, Yin, and Tsang 2023), DiffPool (Ying, R.et al 2018), and ASAP (Ranjan, E.et al 2019). Description and setup of the above models are detailed in the Supplement (section D).

## Implementation Details

The code for this study was implemented in PyTorch Geometric and experimented on WIN10 with a NVIDIA GeForce RTX 3080 Ti GPU with 12GB memory. The model was sequentially set up with 2 convolutional layers, one pooling layer, and finally flattened input to a MLP layer for a fully connected output. After each convolution, batchnorm and a Rectified Linear Unit (ReLU) were used, with MLP hidden layer neurons set to double the number of input features plus one. The study utilized the Adam solver (Kingma and Ba 2014) and binary cross entropy with logits loss, incorporating an initial learning rate of 0.001 and an epoch of 500. Simultaneously, to construct the DSF-BrainNet, we set the window width $W = 95$ and sliding step $s = 10$ for the sliding time window (see Supplement (section A) for details). In COBRE+UCLA dataset, we retained $\gamma = 5$ for the number of subgraphs (clusters), and $\beta n = 3$ for the number of nodes in each subgraph. In REST-meta-MDD dataset, we retained $\gamma = 4$ subgraphs (clusters) and $\beta n = 3$ nodes in each subgraph.

Additionally, all comparative experiments were conducted ten times with five-fold cross-validation, of which four folds were used for training and one fold was used for testing. We then averaged the results from each test set to obtain the final outcome.

## Experiment Results

**Brain SOTA Models.** As shown in Table 2, the proposed model manifests enhanced performance on the two multi-site datasets, surpassing the other baseline methods, thus substantiating its robustness. Specifically, the SVM and GCN yielded the worst results. Among the models utilizing dFC features, specifically SCNN, HFCN, and CRNN, their performance slightly trailed behind that of BrainGNN employing sFC features. In contrast, the proposed EHCPool model exhibited superior performance compared to BrainGNN. It is evident that while dFC holds more unexplored latent feature information than sFC, models employing dFC to exceed the existing SOTA sFC models can only be achieved by utilizing appropriate analytical frameworks and methodologies.

**Graph Pooling Methods.** As shown in Table 3, the proposed method outperforms other node or edge-only graph pooling operators with an average improvement in ACC of 3.44%, 4.30% on the two multi-site datasets, respectively. There are no common graph pooling methods that compute edge scores based on edge features, while EHCPool leverages edge features to compute edge scores. This may explain why EHCPool is superior.

## Ablation Study

To evaluate the effectiveness of different sub-methods employed in the proposed EHCPool model, we executed exhaustive ablation experiments on two mixed datasets.

| Model | Type | COBRE+UCLA (mean (std), %) | | | | REST-meta-MDD (mean (std), %) | | | |
|---|---|---|---|---|---|---|---|---|---|
| | | ACC | SEN | SPE | F1-Score | ACC | SEN | SPE | F1-Score |
| SVM (V. Vapnik 1998) | sFC | 70.62 (4.93) | 69.72 (9.42) | 71.20 (10.07) | 67.05 (9.93) | 72.52 (5.62) | 71.91 (7.25) | 70.42 (9.44) | 69.94 (8.19) |
| GAT-LI (Hu, J. et al. 2021) | sFC | 72.90 (5.85) | 74.70 (9.63) | 72.92 (10.34) | 71.48 (8.91) | 75.38 (6.13) | 73.66 (8.07) | 75.03 (8.56) | 74.61 (9.52) |
| BrainGNN (X. Li et al. 2021) | sFC | 74.11 (7.21) | **75.99** (8.78) | 71.78 (8.64) | 72.62 (6.57) | 77.94 (6.35) | 76.08 (9.54) | **77.18** (8.01) | 75.98 (7.16) |

| Method | | ACC | SEN | SPE | F1-Score | ACC | SEN | SPE | F1-Score |
|---|---|---|---|---|---|---|---|---|---|
| GCN (T. N. Kipf and M. Welling 2016) | dFC | 71.42 (8.48) | 72.13 (9.20) | 70.74 (12.58) | 70.15 (7.92) | 73.86 (7.67) | 71.42 (7.24) | 72.94 (8.66) | 71.25 (10.37) |
| SCNN (Ji, J., Chen, Z., & Yang, C. 2022) | dFC | 72.28 (7.85) | 72.83 (11.73) | 71.15 (10.47) | 71.59 (6.29) | 74.72 (8.34) | 72.37 (11.34) | 74.23 (11.28) | 73.08 (10.68) |
| HFCN (C. Pan et al. 2022) | dFC | 73.84 (7.29) | 70.52 (9.97) | 74.52 (10.82) | 70.48 (9.60) | 76.16 (8.21) | 77.84 (10.29) | 72.46 (11.75) | 74.59 (9.66) |
| CRNN (K. Lin, et al. 2022) | dFC | 72.98 (6.83) | 72.20 (9.02) | 72.83 (10.79) | 72.22 (8.11) | 74.79 (7.34) | 73.72 (8.71) | 73.86 (10.61) | 72.32 (9.57) |
| EHCPool(ours) | dFC | **75.14** (2.62) | **73.57** (4.36) | **75.29** (5.25) | **74.10** (4.31) | **78.83** (3.02) | **79.64** (5.21) | **76.29** (7.64) | **77.51** (4.39) |

Table 2: Comparison of the classification performance with different brain SOTA models

| Method | Type | COBRE+UCLA (mean (std), %) | | | | REST-meta-MDD (mean (std), %) | | | |
|---|---|---|---|---|---|---|---|---|---|
| | | ACC | SEN | SPE | F1-Score | ACC | SEN | SPE | F1-Score |
| Top-kPool (Gao and Ji 2019) | node | 73.68 (2.80) | 72.31 (10.48) | 75.17 (10.72) | 70.29 (8.96) | 75.81 (3.15) | 75.37 (9.49) | 71.04 (11.94) | 73.13 (10.36) |
| SAGPool (Lee, Lee, and Kang 2019) | node | 71.18 (1.81) | 67.06 (5.88) | 74.75 (6.99) | 70.46 (7.79) | 74.47 (2.29) | 72.96 (5.04) | 73.23 (5.67) | 72.87 (7.71) |
| EdgePool (Diehl 2019) | node | 67.56 (3.63) | 65.62 (8.34) | 62.05 (11.27) | 64.19 (11.21) | 70.19 (5.65) | 66.95 (6.31) | 69.54 (10.47) | 68.94 (11.84) |
| MincutPool (Bianchi, Grattarola, and Alippi 2020) | node | 74.36 (3.13) | 72.84 (4.31) | 74.15 (6.72) | 73.42 (7.44) | 77.65 (4.72) | 76.12 (4.56) | 74.36 (5.88) | 74.57 (8.45) |
| Co-Pool (Zhou, Yin, and Tsang 2023) | node | 73.98 (3.78) | 72.75 (5.04) | 73.86 (7.51) | 73.49 (6.93) | 76.75 (5.36) | 75.32 (4.82) | 73.81 (4.56) | 75.43 (9.07) |
| DiffPool (Ying, R.et al 2018) | node | 74.42 (2.29) | 70.68 (5.33) | 74.73 (6.21) | 70.94 (5.17) | 77.78 (3.46) | 76.81 (5.85) | **78.26** (4.91) | 75.46 (6.83) |
| ASAP (Ranjan, E.et al 2019) | node | 70.31 (4.61) | 69.55 (5.82) | 73.61 (8.91) | 66.36 (6.86) | 76.72 (4.37) | 77.52 (5.63) | 74.49 (6.84) | 75.34 (5.76) |
| EHCPool (proposed model) | node+edge | **75.14** (2.62) | **73.57** (4.36) | **75.29** (5.25) | **74.10** (4.31) | **78.83** (3.02) | **79.64** (5.21) | 76.29 (7.64) | **77.51** (4.39) |

Table 3: Classification performance comparison between different graph pooling methods

(i) Scoring approach for node and edge features. Edge-to-Node is a scoring criterion that systematically incorporates both node and edge information, offering a holistic perspective. Therefore, we employed a local scoring strategy using only nodes and a global scoring strategy using only neighboring edges for replacement. (ii) Graph pooling kernel for soft or hard clustering. Owing to the hard clustering design of Iteration n-top, implementing soft clustering pooling within EHCPool by directly replacing Iteration n-top was not feasible. Therefore, two SOTA soft cluster pooling methods, DiffPool and ASAP, were chosen for comparison. The detailed results were shown in Table 3. (iii) Feature aggregation strategy. Two highly representative graph pooling aggregation approaches were substituted for N-E aggregation. Feature Selection is a strategy that selects a portion of node features directly as the output, such as Top-kPool. Fully connected aggregation is a strategy that aggregates the node features within each cluster into the corresponding cluster basis vectors as output, without factoring in the edge feature information, e.g., DiffPool and ASAP.

The results of the first set of experiments, as shown in Table 4, indicate that the Edge-to-Node scoring strategy has better performance than the remaining two scoring strategies. This underscores the importance of integrated global and local representations in graph classification. The second set of experimental results, as shown in Table 3, further validate the superiority of the proposed method over alternative soft clustering graph pooling kernels. This suggests that the repeated representation of critical features in soft clustering may impact the experimental results. The third set of experimental results, as shown in Table 5, indicates that the performance of N-E aggregation surpasses that of the other methods. All the experimental findings mentioned above substantiate the efficacy of each respective sub-method.

| Method | Type | COBRE+UCLA (mean (std), %) | | | | REST-meta-MDD (mean (std), %) | | | |
|---|---|---|---|---|---|---|---|---|---|
| | | ACC | SEN | SPE | F1-Score | ACC | SEN | SPE | F1-Score |
| Node-only features | local | 73.14 (2.91) | 72.47 (7.94) | 72.83 (8.76) | 70.30 (4.19) | 76.41 (4.27) | 75.44 (6.86) | 75.03 (7.35) | 75.62 (5.08) |
| Edge-only features | global | 74.36 (5.61) | **74.57** (6.52) | 73.22 (5.01) | 72.34 (5.47) | 77.56 (4.83) | 78.81 (5.09) | 75.74 (4.36) | 75.28 (5.65) |
| Edge-to-Node score strategy(ours) | local+global | **75.14** (2.62) | 73.57 (4.36) | **75.29** (5.25) | **74.10** (4.31) | **78.83** (3.02) | **79.64** (5.21) | **76.29** (7.64) | **77.51** (4.39) |

Table 4: Comparison between different feature scoring strategies in pooling kernel

| Method | Type | COBRE+UCLA (mean (std), %) | REST-meta-MDD (mean (std), %) |
|---|---|---|---|

|  |  | ACC | SEN | SPE | F1-Score | ACC | SEN | SPE | F1-Score |
|---|---|---|---|---|---|---|---|---|---|
| Feature selection | node | 70.29 (4.28) | 71.25 (6.46) | 70.40 (5.88) | 69.82 (4.10) | 74.15 (5.58) | 72.61 (6.65) | 74.33 (8.62) | 73.06 (5.35) |
| Fully Connected aggregation | node | 74.84 (2.14) | **74.52** (7.73) | 72.01 (8.59) | 73.46 (5.95) | 76.97 (3.71) | 77.74 (6.71) | 75.75 (6.86) | 76.46 (5.66) |
| N-E Aggregation (ours) | node+edge | **75.14** (2.62) | 73.57 (4.36) | **75.29** (5.25) | **74.10** (4.31) | **78.83** (3.02) | **79.64** (5.21) | **76.29** (7.64) | **77.51** (4.39) |

Table 5: Comparison between different information aggregation strategies in pooling kernel

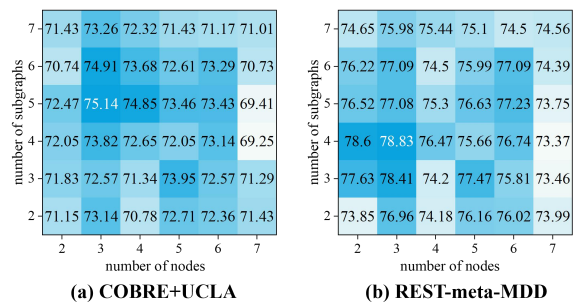

Figure 3: Discussion of $\gamma$ and $\beta$ parameters

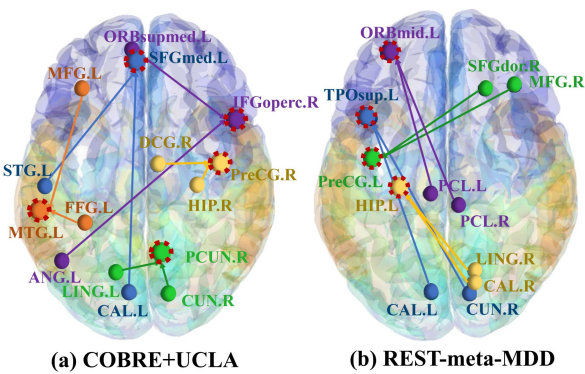

Figure 4: Proposed model subgraph division results and visualization at the highest accuracy. Different colors represent different subgraphs, outline highlighted nodes are core nodes. Full names of different brain regions are detailed in the Supplement (section E).

### Sensitivity Analysis and Visualization

A sensitivity analysis was performed on the number of subgraphs (clusters) to be retained $\gamma$ and the ratio of retained nodes per subgraph $\beta$. As shown in Figure 3, the model ac curacy is sensitive to these two hyperparameters, indicates that multiple different types of abnormal functional brain networks (composed of different brain regions) may exist in psychiatric patients. It is highly challenging to investigate various abnormal functional brain networks related to psychiatric disorders solely from a data-driven perspective without prior knowledge. To the best of our knowledge, there is currently shortage of published deep learning investigations addressing this issue. Of all the studies related to this topic, only BNC-DGHL (Ji and Zhang 2022) has simply combined the brain regions corresponding to the top 10 most abnormal FCs, treating them as a whole-brain abnormal functional network rather than exploring multiple different types of abnormal functional networks within the brain. However, the results of the sensitivity analyses, as portrayed in Figure 4, suggest that EHCPool can differentiate and categorize abnormal brain regions into different abnormal subgraphs, and thus to identify abnormal subnetworks in the brain that represent different types of functional abnormalities. For further discussion, refer to the Supplement (Section F).

### Conclusion

This study proposes EHCPool, the first graph clustering pooling method designed around edge features and innovates the graph clustering process. Specifically, the 'Edge-to-Node' score criterion for evaluating node significance based on edge features is presented, which improves the graph pooling representation learning capability. Subsequently, Iteration n-top captures the critical subgraphs through hard clustering dominated by edge features, which further exploits the application prospects of graph pooling methods and is expected to significantly improve interpretability of the model in specific scenarios. Finally, the proposed N-E Aggregation conducts an independent information aggregation in each subgraph, which provides a novel pooling aggregation concept for graph structures with both node and edge features. The effectiveness of EHCPool was validated in terms of its classification performance on two multi-site brain imaging public datasets. More meaningfully, in the realm of deep learning research focused on medical imaging, EHCPool serves as a potential method to explore different types of abnormal functional brain networks. Future work will concentrate on constructing specialized datasets to probe abnormal functional brain networks in psychiatric patients using EHCPool.